\pdfoutput=1

\documentclass[11pt]{article}

\usepackage{acl}
\usepackage{subcaption}
\usepackage{hyperref}

\usepackage{times}
\usepackage{latexsym}

\usepackage[T1]{fontenc}
\usepackage{mdframed}
\usepackage{booktabs} 
\usepackage{siunitx}  

\usepackage[utf8]{inputenc}
\usepackage{todonotes}
\usepackage{microtype}
\usepackage{graphicx}
\usepackage[toc,page]{appendix}
\usepackage[most]{tcolorbox}
\usepackage{caption}

\tcbuselibrary{skins,breakable}

\newcommand{\System}[1]{\textbf{System:} #1\par}
\newcommand{\SystemRole}[1]{\textbf{System Role:} #1\par}
\newcommand{\Assistant}[1]{\textbf{Assistant:} #1\par}
\newcommand{\User}[1]{\textbf{User:} #1\par}

\title{ChatGPT as a Solver and Grader \\ of Programming Exams written in Spanish}

\author{Pablo Saborido-Fernández, Marcos Fernández-Pichel,\\ \bf{\and David E. Losada}\\
  Centro Singular de Investigación en Tecnoloxias Intelixentes (CiTIUS), \\Universidade de Santiago de Compostela, Spain \\
  {\tt pablo.saborido@rai.usc.es,} \\{\tt \{marcosfernandez.pichel,david.losada\}@usc.es}}

\begin{document}
{\makeatletter\acl@finalcopytrue
  \maketitle
} 

\begin{abstract}
Evaluating the capabilities of Large Language Models (LLMs) to assist teachers and students in educational tasks is receiving increasing attention. In this paper, we assess ChatGPT's capacities to solve and 
grade real programming exams, from an accredited BSc degree in Computer Science, written in Spanish. Our findings suggest that this AI model is only effective for solving simple coding tasks. Its proficiency in tackling complex problems or evaluating solutions authored by others are far from effective. 
As part of this research, we also release a new corpus of programming tasks and the corresponding prompts for solving the problems or grading the solutions. This resource can be further exploited by other research teams. 
\end{abstract}


\section{Introduction}

Large Language Models (LLMs) 
based on the Transformer architecture~\cite{vaswani2017attention}
have represented a paradigm shift in Natural Language Processing (NLP), making the previous state-of-the-art results and benchmarks obsolete. 
The release of ChatGPT 
by OpenAI in November 2022 meant a disruption in what was thought to be possible in generating human-like conversations~\cite{chatgpt}. This class of generative models has proved to be effective in a wide range of 
Natural Language Processing (NLP) tasks~\cite{zhong2023can,mao2023gpteval}.


These new tools have also demonstrated impressive capabilities for solving programming tasks. This is often attributed to the fact that their internal models have been exposed to a large number of programming examples during their training process \cite{zhong2023can,xu2022systematic,chen2021evaluating}. LLMs can thus become a highly valuable asset to support 
different teaching activities in multiple computer-related courses and university degrees~\cite{baidoo2023education}. 
For example, we 
can exploit them to give support
to students in problem solving, to suggest exercises and activities to the professors, or to assist in the grading processes. But this also comes with drawbacks, such as those related to plagiarism or cheating.

Some researchers have already tested ChatGPT for 
resolving programming problems, demonstrating human-level performance for simple tasks, but also finding that it struggles with complex data structures~\cite{chang2023survey,sarsa2022automatic}. However, most studies and benchmarks have been confined to exams written in English. 
Although the multilingual settings of many LLMs have made it possible to apply these models to other languages, the performance is often lower than that of English and
more scientific efforts are needed to evaluate the benefits 
and limitations of these language models for other languages.

Furthermore, most previous studies focused on resolving simple programming tasks, such as basic coding exercises. Our evaluation addresses not only basic programming challenges, but also more complex exercises that require
reasoning about computational complexity, 
decision making about algorithmic strategies, or selecting proper data structures (e.g., stacks or queues). 

Another aspect that has received little attention is the role of LLMs as graders or assistants in evaluating the quality of solutions written by humans. By advancing
our understanding on the 
grading abilities of AI agents, we can shed light on the feasibility of incorporating them into new (student-machine) learning activities or even  exploiting them to automatically or 
semi-automatically grade academic assignments. 

In this study, we evaluate ChatGPT's abilities to solve programming and algorithmic problems extracted 
from a real exam written in Spanish. 
The exam, which is the final test of a 1st-year/2nd-semester course on Programming within a BSc in Computer Science, covers a wide range of exercises, from basic coding exercises to more intricate reasoning tasks. We also assess here the AI's capacities to evaluate exams solved by students enrolled in the course from which the exam was taken. 


Therefore, our contributions are:

\begin{itemize}
    \item An evaluation of how well ChatGPT solves complex programming and algorithmic problems written in Spanish.
    \item A study of the feasibility of ChatGPT to act as an automatic evaluator for tests solved by university students.
    \item A detailed item-by-item analysis
    of the strengths and weaknesses of ChatGPT as a solver and grader of multiple programming exercises.
    \item A new corpus of programming tasks
    and the corresponding set of prompts to ask the models to solve problems or grade solutions. This new resource can be further exploited by other research teams to conduct further evaluations of LLMs for programming problem resolution. All the data and code of this research is freely available for the scientific community\footnote{\url{https://anonymous.4open.science/r/chatgpt-grader-solver-exams-7BBF/}}.
    
\end{itemize}




\section{Method}

\subsection{ChatGPT as a Solver}

We chose a real exam from a
1st year-2nd semester course on Programming, Linear Data Structures and Introduction to Computational Complexity. The exam was taken in May 2023 by 90 students from an 
accredited BSc degree in Computer
Science. The average score of
the students was 57.55\% (std dev 20.29\%), 26 of them did not pass (score below 50\%), and 5 students scored above 90\%. 

 It should be noted that this is an exam that tests not only basic coding skills, but also algorithmic and data structure concepts. The exam consisted of 7 questions (see Table \ref{t1a}, Appendix~\ref{app:b}), with varied types of expected responses, ranging from a full page to a short textual answer. Some of the questions involved the development of C code. In the original exam, two questions (\#3 and \#5) had two figures that further help to clarify the particular inquiry. Since ChatGPT\footnote{We used gpt-3.5-turbo version for this experimentation} does not accept images, we opted for removing the images. In any case, the images were redundant (e.g. one represented the internal structure of the Abstract Data Type (ADT) list, whose code was given in the text of the question) and one can understand the question without having to see the image. 
 However, we have to bear in mind that this may be a small disadvantage for the AI model. The evaluation of more advanced models, such as GPT-4 \cite{gpt4}, was left for future work.


 
Each question was passed to ChatGPT through OpenAI's Python API. Two different prompt variants were tested: a simple one with almost no context (\textbf{Simple Prompt}) and a more sophisticated prompt including formatting instructions and system role (\textbf{Complex Prompt}), see Appendix~\ref{app:a1}. The answers outputted by the AI model were given to the main instructor of the course (a professor in CS\&AI), who assessed the model's solutions using the same criteria set for the official exam. 

\subsection{ChatGPT as a Grader}

We also wanted to evaluate ChatGPT's capacity to assess the quality of human-made solutions. 
The official exams solved by the students
were manuscript and, thus, we can hardly evaluate them all. Instead, we chose a sample of five exams, with a varied range of scores (94\%, 74\%, 66\%, 50\% and 38\%), transcribed 
them and submitted them to the model's API. 
We sent each question individually and asked the model to provide a quality score (0\%-100\%), see Appendix~\ref{app:a2}. Then, an overall grade was obtained by weighting the questions using the point scale established in the official exam.


\section{Results}

\subsection{ChatGPT as a Solver} \label{sec:solver}

\begin{table*}[t!]
\centering
\footnotesize
\sisetup{
  table-format=1.2,
  table-space-text-post={/2}, 
}
\begin{tabular}{
  l
  S[table-format=1.2]
  S[table-format=1.2]
  S[table-format=1.2]
  S[table-format=1.2]
  S[table-format=1.2]
  S[table-format=1.2]
  S[table-format=1.2]
  c
}
\toprule
& \multicolumn{7}{c}{Questions} & {Overall Grade} \\
\cmidrule(lr){2-8} \cmidrule(lr){9-9}
& {1} & {2} & {3} & {4} & {5} & {6} & {7} & \\
\midrule
Points & 1.5 & 1.5 & 1.5 & 1 & 1 & 2 & 1.5 &  \\
Simple Prompt  & {0\%} & {83.3\%} & {66.7\%} & {100\%} & {100\%} & {100\%} & {16.7\%} & \textbf{65\%} \\
Complex Prompt & {0\%} & {43.3\%} & {66.7\%} & {100\%} & {100\%} & {75\%} & {0\%} & \textbf{51.5\%} \\
\bottomrule
\end{tabular}
\caption{Grades obtained by ChatGPT. The first row shows the maximum number of points per question}
\label{tab:results-solving}
\end{table*}

Table~\ref{tab:results-solving} shows the results achieved by ChatGPT in the exam. As can be seen in the first row, each question had a different number of points. 
To avoid any possible bias, the professor did not know which prompt generated each version of the responses. 

The first noticeable result is that, for both variants, the model achieved a score above the required threshold to pass the final exam.
This is not a minor outcome, since previous research has demonstrated that these models struggle with difficult data structure tasks~\cite{chang2023survey}. ChatGPT's grades are similar to those achieved by the average
student. One might argue that matching human
performance is profoundly meaningful. However,
we see here two main sources of concern.
First, the students examined are 
novice undergraduates in the first year of training (most of them with only a few months of experience in programming). So, ChatGPT is not really matching expert-level performance. 
Second, ChatGPT's performance does not
place it in a position to be utilised as an 
intelligent assistant. You can hardly exploit
a tool that produces wrong results more than 30\% of the time. 

A second interesting result is that the use of a complex prompt
did not help the model. The complex prompt 
was never better than the simple prompt.
No single question got a better response from the complex prompt. 
The specification of
system role, the ``take-your-time'' advise
and the provided example
do not seem to be useful and, 
perhaps, have introduced some confusion.

Regarding specific questions, both 
variants struggled with question number 1  (syntactic and semantic specification of an ADT) and number 7 (reasoning about an example of
divide and conquer algorithm). In question 1, 
ChatGPT did not output a formal specification
of the ADT, failed to provide a semantic description with proper algebraic notation
and often resorted to not-allowed expressions
(e.g. using integer expressions in C rather than
generic numerical expressions). We conjecture
that this might be related to the low availability of ADT examples with proper notations in the training data. Question \#7 was
about interpreting different levels of computational complexity of a divide and conquer solution, based on variables such as the number of subproblems, size of the subproblems and so forth. This type of conceptual question also made that ChatGPT failed loudly. For the
rest of the questions (\#2-\#6) ChatGPT made
a decent job. Three of them were mainly coding tasks (\#3, \#5, \#6) and two of them (\#2, \#4)
required some sort of reasoning but they refer
to well-known computing examples (Fibonacci or list search). 
In some instances, the model did not follow the instructions and, rather than outputting solutions
written in C, it provided correct solutions written in Python. 
Note also that ChatGPT did well
on questions \#3 and \#5, which had a supporting image that the model could not see.



\subsection{ChatGPT as a Grader} 


\begin{figure}[t!]
\centering
\includegraphics[width=1.1\columnwidth]{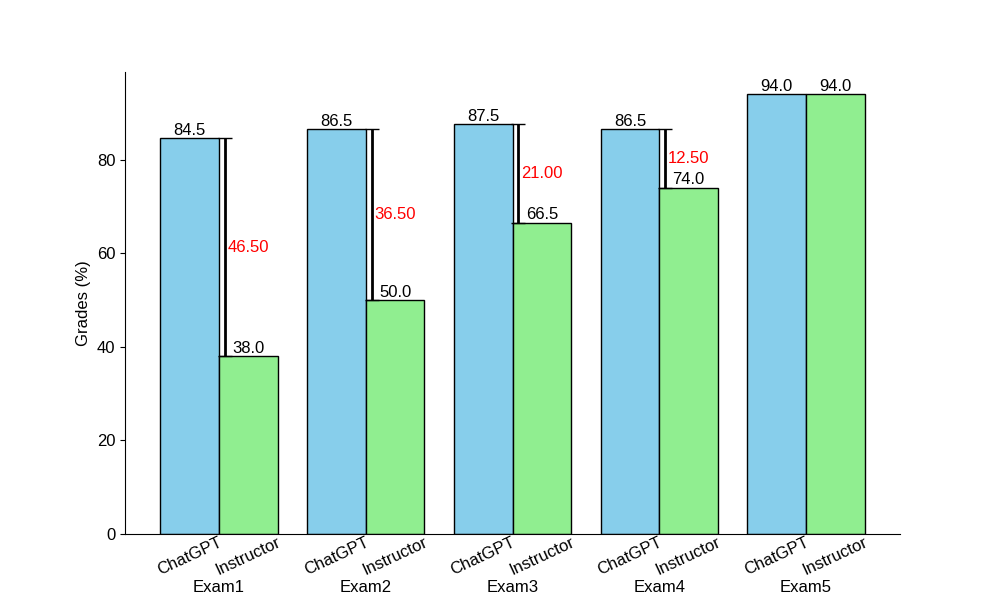}
\caption{\label{fig:chat_human_eval} ChatGPT as a grader. For each exam solved by a student, 
the bars represent the score given by ChatGPT
and the score given by the instructor of the course.} 
\end{figure}



Figure~\ref{fig:chat_human_eval} shows the grades assigned by ChatGPT and by the course's instructor to the five exams
selected. The AI model clearly overestimates the quality of the solutions and, indeed, all exams got a high qualification (all of them above 84\%). Even low quality solutions, such
as the exam that officially got a 38\% overall score, were assigned very good scores. This hardly positions ChatGPT as a tool to assist humans (professors or students) in the assessment of solutions for this type of exams. 

Next, we analyse the individual question-by-question
assessments, see Table~\ref{tab:comparison-grades} in Appendix~\ref{app:c}. 
We report the scores assigned and 
the deviation between ChatGPT and the instructor. 
The largest deviation was found in Question 1, on ADT 
specification. This result is in agreement with the findings in Section~\ref{sec:solver}, in which the model also struggled to solve this exercise. Again, this suggests that the model has little knowledge about this topic or 
it is not able to transfer its knowledge to produce answers
that comply with the instructions.
The only decent grading by ChatGPT was done for question 4.
This question was effectively solved by ChatGPT (see 
section \ref{sec:solver}) and, here, the model also shows
reasonably good performance at evaluating question 4's answers written by students. 
These answers are short paragraphs explaining
the computational complexity of a given search problem (traversing over a list). A somehow surprising result
is that ChatGPT was highly effective at producing solutions
for Question 5 but it drastically failed to assess
the quality of Question 5 solutions written by students. 
This was a C function that implements a recursive process and ChatGPT was unable to effectively assess the quality of the functions written by students. Furthermore, ChatGPT's tendency
to overrate the quality of the solutions was consistent
over all types of questions\footnote{As a side note, we observe that most students would have loved to have ChatGPT as grader, replacing the official instructor.}.


Note also that ChatGPT assigned an overall score to
the best exam that was the same score assigned by
the instructor (94\%). But this seems to be anecdotal, as the individual question-by-question 
scores (Exam5) show substantial deviations. 


\section{Discussion}

The results of this study
suggest that ChatGPT performs much better 
on solving exercises than it does on
grading them. 
As a solver, it is worth noting the AI model's poor performance in some types of exercises (particularly in those that 
do not involve coding). 
In the near future, it will be beneficial to investigate the reasons for this poor performance. For instance, to understand whether it is related to the language 
(Spanish) in which the problems are expressed or due to a lack of training data for these types of exercises.
The specific phrasing might have also played
an important role, as ChatGPT
performed poorly for exercise 7, while other questions --also about computational complexity-- had much better answers
from the model. 
This suggests that wording might have a strong influence on model's performance.

On the other hand, our results suggest that the model is useless to validate  answers submitted by humans. Even with coding questions that the model solved very well, its assessment of solutions written by others was unsatisfactory. 



\section{Related Work}

The development of Large Language Models (LLMs) has represented a paradigm shift for multiple NLP tasks \cite{brown2020language,kojima2022large}. In some cases, these models even reach human performance. For instance, previous studies evaluated GPT-4, demonstrating its ability for several reasoning-intensive tests, such as passing a technical entrance exam for a software engineering position \cite{bubeck2023sparks}.

The ability of these models to generate code has sparked interest in both the developer and teaching communities. In this direction, Chen et al.~\cite{chen2021evaluating} introduced Codex, a model specifically fine-tuned for solving programming tasks. Other authors studied Codex's potential for solving Python tasks, demonstrating its effectiveness on the APPS benchmark \cite{hendrycks2021measuring}. 
The authors of \cite{sarsa2022automatic} explored the possibility of integrating Codex in teaching duties, for example to generate coding exercises. Similarly, Xu and colleages \cite{xu2022systematic} conducted a systematic evaluation of different language models --both proprietary like Codex and open source-- for coding completion and synthesis tasks. These authors also proposed a novel fine-tuned model that outperforms other alternatives for C programming. Chang et al. \cite{chang2023survey} stated that ChatGPT outperforms humans in simple coding assignments, but still struggles with data structure problems and graph theory. 

In this paper, we have presented a novel approach to evaluate ChatGPT's programming abilities. We evaluated its capacities to solve assorted programming-related exercises, spanning multiple areas
such as abstract data types, data structures and computational complexity.  Another novelty that distinguishes our work from previous contributions is that 
we also studied the AI model's
capacity as a grader. 
Furthermore, our study targeted
the Spanish language, thus responding to the growing interest of the scientific community in evaluating the capabilities of LLMs in languages other than English \cite{deng2023multilingual}.

\section{Conclusion}

In this study, we assessed ChatGPT's capacities to solve and grade 
programming exams (in Spanish) from an
official university course. The results suggest that this AI model can only be used as a solver of basic coding exercises. Its abilities to solve and reason about intricate questions, and its capacity to assess solutions written by others are far from effective. The study of more sophisticated prompting strategies, such as those based on paraphrasing the original instructions, are left as future work.

\newpage

\section*{Ethics Statement and Limitations}

This research aims at 
evaluating the capabilities of the new generative AI models as a support tool in educational environments. Our study 
was constrained to exams written in Spanish because we were
specifically interested in
analysing how LLMs perform in languages other than English. 
Access to the exams was provided to us by the main instructor of the university course. The exams filled by the students were anonymised and their responses to the questions did not contain any personal or private reference.

The overall goal of this project  is to gain an understanding  on how the new AIs could help 
to automate or ease certain learning and grading tasks. This research does not pursue 
the elimination of human instructors from the university classes. As a matter of fact, we
firmly believe that human-in-the-loop strategies are crucial to properly exploit the advantages and reduce the risks of AI-based
agents.

We are also aware that
more sophisticated models, such as GPT4, could perform better. But, currently, ChatGPT is a model that is freely available
and it already has a 
huge user base worldwide
(including many university students). Thus, our study was centered on the most popular 
and widely available
platform. Anyway, in the future we will extend this research to other LLMs. 
We also recognise that more sophisticated prompt engineering could lead to better performance. We left this exploration for future research and we decided to employ here two initial types of prompts.


\bibliography{acl_latex}

\begin{thebibliography}{15}
\expandafter\ifx\csname natexlab\endcsname\relax\def\natexlab#1{#1}\fi

\bibitem[{Baidoo-Anu and Ansah(2023)}]{baidoo2023education}
David Baidoo-Anu and Leticia~Owusu Ansah. 2023.
\newblock Education in the era of generative artificial intelligence (ai): Understanding the potential benefits of chatgpt in promoting teaching and learning.
\newblock \emph{Journal of AI}, 7(1):52--62.

\bibitem[{Brown et~al.(2020)Brown, Mann, Ryder, Subbiah, Kaplan, Dhariwal, Neelakantan, Shyam, Sastry, Askell et~al.}]{brown2020language}
Tom Brown, Benjamin Mann, Nick Ryder, Melanie Subbiah, Jared~D Kaplan, Prafulla Dhariwal, Arvind Neelakantan, Pranav Shyam, Girish Sastry, Amanda Askell, et~al. 2020.
\newblock Language models are few-shot learners.
\newblock \emph{Advances in neural information processing systems}, 33:1877--1901.

\bibitem[{Bubeck et~al.(2023)Bubeck, Chandrasekaran, Eldan, Gehrke, Horvitz, Kamar, Lee, Lee, Li, Lundberg et~al.}]{bubeck2023sparks}
S{\'e}bastien Bubeck, Varun Chandrasekaran, Ronen Eldan, Johannes Gehrke, Eric Horvitz, Ece Kamar, Peter Lee, Yin~Tat Lee, Yuanzhi Li, Scott Lundberg, et~al. 2023.
\newblock Sparks of artificial general intelligence: Early experiments with gpt-4.
\newblock \emph{arXiv preprint arXiv:2303.12712}.

\bibitem[{Chang et~al.(2023)Chang, Wang, Wang, Wu, Zhu, Chen, Yang, Yi, Wang, Wang et~al.}]{chang2023survey}
Yupeng Chang, Xu~Wang, Jindong Wang, Yuan Wu, Kaijie Zhu, Hao Chen, Linyi Yang, Xiaoyuan Yi, Cunxiang Wang, Yidong Wang, et~al. 2023.
\newblock A survey on evaluation of large language models.
\newblock \emph{arXiv preprint arXiv:2307.03109}.

\bibitem[{Chen et~al.(2021)Chen, Tworek, Jun, Yuan, Pinto, Kaplan, Edwards, Burda, Joseph, Brockman et~al.}]{chen2021evaluating}
Mark Chen, Jerry Tworek, Heewoo Jun, Qiming Yuan, Henrique Ponde de~Oliveira Pinto, Jared Kaplan, Harri Edwards, Yuri Burda, Nicholas Joseph, Greg Brockman, et~al. 2021.
\newblock Evaluating large language models trained on code.
\newblock \emph{arXiv preprint arXiv:2107.03374}.

\bibitem[{Deng et~al.(2023)Deng, Zhang, Pan, and Bing}]{deng2023multilingual}
Yue Deng, Wenxuan Zhang, Sinno~Jialin Pan, and Lidong Bing. 2023.
\newblock Multilingual jailbreak challenges in large language models.
\newblock \emph{arXiv preprint arXiv:2310.06474}.

\bibitem[{Forbes(2022)}]{chatgpt}
Forbes. 2022.
\newblock \href {https://openai.com/blog/chatgpt} {Introducing chatgpt}.
\newblock [accessed April 4, 2023].

\bibitem[{Hendrycks et~al.(2021)Hendrycks, Basart, Kadavath, Mazeika, Arora, Guo, Burns, Puranik, He, Song et~al.}]{hendrycks2021measuring}
Dan Hendrycks, Steven Basart, Saurav Kadavath, Mantas Mazeika, Akul Arora, Ethan Guo, Collin Burns, Samir Puranik, Horace He, Dawn Song, et~al. 2021.
\newblock Measuring coding challenge competence with apps.
\newblock \emph{arXiv preprint arXiv:2105.09938}.

\bibitem[{Kojima et~al.(2022)Kojima, Gu, Reid, Matsuo, and Iwasawa}]{kojima2022large}
Takeshi Kojima, Shixiang~Shane Gu, Machel Reid, Yutaka Matsuo, and Yusuke Iwasawa. 2022.
\newblock Large language models are zero-shot reasoners.
\newblock \emph{Advances in neural information processing systems}, 35:22199--22213.

\bibitem[{Mao et~al.(2023)Mao, Chen, Zhang, Guerin, and Cambria}]{mao2023gpteval}
Rui Mao, Guanyi Chen, Xulang Zhang, Frank Guerin, and Erik Cambria. 2023.
\newblock Gpteval: A survey on assessments of chatgpt and gpt-4.
\newblock \emph{arXiv preprint arXiv:2308.12488}.

\bibitem[{OpenAI(2023)}]{gpt4}
OpenAI. 2023.
\newblock Gpt-4 technical report.
\newblock \emph{arXiv:submit/4812508}.

\bibitem[{Sarsa et~al.(2022)Sarsa, Denny, Hellas, and Leinonen}]{sarsa2022automatic}
Sami Sarsa, Paul Denny, Arto Hellas, and Juho Leinonen. 2022.
\newblock Automatic generation of programming exercises and code explanations using large language models.
\newblock In \emph{Proceedings of the 2022 ACM Conference on International Computing Education Research-Volume 1}, pages 27--43.

\bibitem[{Vaswani et~al.(2017)Vaswani, Shazeer, Parmar, Uszkoreit, Jones, Gomez, Kaiser, and Polosukhin}]{vaswani2017attention}
Ashish Vaswani, Noam Shazeer, Niki Parmar, Jakob Uszkoreit, Llion Jones, Aidan~N Gomez, {\L}ukasz Kaiser, and Illia Polosukhin. 2017.
\newblock Attention is all you need.
\newblock \emph{Advances in neural information processing systems}, 30.

\bibitem[{Xu et~al.(2022)Xu, Alon, Neubig, and Hellendoorn}]{xu2022systematic}
Frank~F Xu, Uri Alon, Graham Neubig, and Vincent~Josua Hellendoorn. 2022.
\newblock A systematic evaluation of large language models of code.
\newblock In \emph{Proceedings of the 6th ACM SIGPLAN International Symposium on Machine Programming}, pages 1--10.

\bibitem[{Zhong et~al.(2023)Zhong, Ding, Liu, Du, and Tao}]{zhong2023can}
Qihuang Zhong, Liang Ding, Juhua Liu, Bo~Du, and Dacheng Tao. 2023.
\newblock Can chatgpt understand too? a comparative study on chatgpt and fine-tuned bert.
\newblock \emph{arXiv preprint arXiv:2302.10198}.

\end{thebibliography}

\newpage

\begin{appendices}

\section{Prompts} \label{app:a}
\subsection{Prompts for Solving} \label{app:a1}

\subsubsection{Simple Prompt}

Under this setting, no further instruction but the exam question was provided to the model:

\begin{mdframed}
\User{\textit{<QUESTION>}}
\System{...}
\end{mdframed}

\subsubsection{Complex Prompt}

Under this setting, we specified a system role to give more context to the model about the task. 
Additionally, the large prompt specifies that the model can take ``time to reason'' (this is a recommended practice in these tasks)
and, additionally, 
we also provided a demonstration, which consists of a natural language instruction (asking to build a hello world program) and
the corresponding C code.

\begin{mdframed}
\SystemRole{Estás respondiendo a las preguntas de un examen de informática centrado en el lenguaje de programación C.}
\User{Escribe un programa en C que escriba: Hello World}
\Assistant{El siguiente código está escrito en C:\\\#include <stdio.h>\\int main(void)\\\{\\printf("Hello World");\\return 0;\\\}}
\User{La siguiente es una pregunta de un examen de programación del primer año del grado de ingeniería informática. Hay bastante tiempo para responder, así que tómate el tiempo que sea necesario para dar una respuesta completa y razonada paso a paso. La pregunta está delimitada por < >. Además en el caso de que tengas que escribir código, primero especifica el lenguaje en el que está escrito, y luego escribe dicho código delimitándolo con ''' antes de la primera línea y después de la última, tal y como has hecho anteriormente:\\<\textit{QUESTION}>}
\System{...}
\end{mdframed}

\subsection{Prompt for Grading} \label{app:a2}

Under this setting, the prompt
asks the model to reason about
the model's response to the question and, next, it asks the model to compare it against the provided response and, finally, give an overall quality score for the provided response. 

 \begin{mdframed}
\User{Tu tarea es evaluar la respuesta a una pregunta de un examen de programación. Para ello razona primero tu respuesta y compárala con la respuesta proporcionada. No la evalúes hasta que no hayas respondido tú mismo a la pregunta. La pregunta está delimitada por <...> y la respuesta a evaluar está entre "...". El formato de tu respuesta debe ser el siguiente, respétalo sin añadir ningún comentario adicional y asegúrate de escribir una nota numérica sobre 100: \\ Pregunta: (copia aquí la pregunta del examen entre <...>) \\ Respuesta: (copia aquí la respuesta del alumno entre "...") \\ Calificación: La nota es (nota sobre 100\%)\\<\textit{QUESTION}>\\``\textit{RESPONSE}''}
\System{...}
\end{mdframed}

\section{Exam Questions}
\label{app:b}

Table~\ref{t1a} details the questions that made up the exam.

\begin{table*}[h!]
\footnotesize
\center
\begin{tabular}{clc}
{\bf Question} & {\bf Topic} & {\bf Response Type} \\ \hline
\#1 & Syntactic and semantic formal specification of & Full page \\
 & an abstract data type (ADT) &  \\
\#2 & Recursive implementation of Fibonacci. Reason about...  & 3 short responses  \\
& i. worst/best/avg time complexity
& (1-2 sentences each) \\
& ii. type of algorithmic strategy
&  \\
& iii. \# invoked instances
&  \\
\#3 & ADT List (internal data structure is given) & 2 short C functions  \\
& i. \& ii. implement functions get/create
& + 2 one-word responses \\
& iii. \& iv. worst case time complexity of functions get/create
&  \\
\#4 & List. Worst/Best Time complexity of a given search problem & One paragraph \\
\#5 & Implementation in C of a recursive function that solves & Short C function \\
 & a given problem &  \\
 \#6 & Implementation in C of a function that uses an ADT Queue & C function \\
 & (with provided operations) to solve a given problem &  \\
 \#7 & Divide \& Conquer Example. Reason about... & 3 short responses \\
 & i. parameters (\# subproblems, split/aggregation costs, & (2-3 sentences each) \\
  & subproblem sizes) & \\
  & ii. compare two variants for the same problem & \\
  & iii. compare against 2 types of sequential search & \\ \hline
\end{tabular}
\caption{Exam Questions}
\label{t1a}
\end{table*}

\section{Evaluating Results}
\label{app:c}

Table~\ref{tab:comparison-grades} breaks down the grades assigned by both assessors.

\begin{table*}[t!]
\centering
\scriptsize
\begin{tabular}{cccc|ccc|ccc|ccc}
\cmidrule{2-13}
 & \multicolumn{3}{c|}{Question 1} & \multicolumn{3}{c|}{Question 2} & \multicolumn{3}{c|}{Question 3} & \multicolumn{3}{c}{Question 4} \\
  & {ChatGPT} & {Instructor} & {Dev} & {ChatGPT} & {Instructor} & {Dev} & {ChatGPT} & {Instructor} & {Dev} & {ChatGPT} & {Instructor} & {Dev} \\
\midrule
 Exam1 & 100\% & 53.3\% & +46.7\% & 100\% & 50\% & +50\% & 50\% & 0\% & +50\% & 90\% & 50\% &  +40\% \\
 Exam2 & 100\% & 16.7\% & +83.3\% & 100\% & 33.3\% & +66.7\% & 25\% & 66.7\% & -41.7\% & 100\% & 100\% &  0\% \\
 Exam3 & 100\% & 50\% & +50\% & 66.7\% & 83.3\% & -16.7\% & 75\% & 60\% & +15\% & 100\% & 100\% &  0\% \\
 Exam4 & 100\% & 16.7\% & +73.3\% & 100\% & 83.3\% & +16.7\% & 25\% & 33.3\% & -8.8\% & 100\% & 100\% &  0\% \\
 Exam5 & 100\% & 100\% & 0\% & 93.3\% & 100\% & -6.7\% & 66.7\% & 100\% & -33.3\% & 100\% & 100\% &  0\% \\ \hline
 Avg. & 100\% & 47.3\% & 50.7\% & 92\% & 70\% & 31.4\% & 48.3\% & 52\% & 29.8\% & 98\% & 90\% &  8\% \\
\bottomrule
\end{tabular}
\end{table*}

\begin{table*}[t!]
\scriptsize
\centering
\begin{tabular}{cccc|ccc|ccc}
\cmidrule{2-10}
 & \multicolumn{3}{c|}{Question 5} & \multicolumn{3}{c|}{Question 6} & \multicolumn{3}{c}{Question 7} \\
  & {ChatGPT} & {Instructor} & {Dev} & {ChatGPT} & {Instructor} & {Dev} & {ChatGPT} & {Instructor} & {Dev} \\
\midrule
 Exam1 & 100\% & 0\% & +100\% & 90\% & 50\% & +40\% & 66.7\% & 50\% & +16.7\% \\
 Exam2 & 100\% & 100\% & 0\% & 90\% & 50\% & +40\% & 90\% & 16.7\% & +73.3\%  \\
 Exam3 & 90\% & 0\% & +90\% & 80\% & 75\% & +5\% & 100\% & 83.3\% & +16.7\% \\
 Exam4  & 100\% & 100\% & 0\% & 100\% & 100\% & 0\% & 93.3\% & 93.3\% & 0\% \\
 Exam5 & 100\% & 100\% & 0\% & 100\% & 75\% & +25\% & 100\% & 93.3\% & +6.7\% \\ \hline
 Avg. & 98\% & 60\% & 38\% & 92\% & 70\% & 22\% & 90\% & 67.3\% & 22.7\% \\
\bottomrule
\end{tabular}
\caption{Comparison of scores assigned by ChatGPT and by the course's instructor.}
\label{tab:comparison-grades}
\end{table*}

\end{appendices}




\end{document}